# Prediction Using Note Text
## Synthetic Feature Creation with *word2vec*

Manuel Amunategui (Manuel.Amunategui@providence.org)
Tristan Markwell (Tristan.Markwell@providence.org)
Yelena Rozenfeld (Yelena.Rozenfeld@providence.org)

## Abstract

*word2vec* affords a simple yet powerful approach of extracting quantitative variables from unstructured textual data. Over half of healthcare data is unstructured [1] and therefore hard to model without involved expertise in data engineering and natural language processing. *word2vec* can serve as a bridge to quickly gather intelligence from such data sources.

In this study, we ran 650 megabytes of unstructured, medical chart notes from the Providence Health & Services electronic medical record through *word2vec*. We used two different approaches in creating predictive variables and tested them on the risk of readmission for patients with COPD (Chronic Obstructive Lung Disease). As a comparative benchmark, we ran the same test using the LACE risk model [2] (a single score based on length of stay, acuity, comorbid conditions, and emergency department visits).

Using only free text and mathematical might, we found *word2vec* comparable to LACE in predicting the risk of readmission of COPD patients.

## Introduction

### Predicting COPD Readmission
This study uses COPD patient data from Providence Health & Services for the discharges from 2012 through 2014, in an attempt to predict whether or not the patient was readmitted within 30 days of discharge. We used the months of 01/2012 to 06/2014 for model training and 07/2014 to 12/2014, for model testing.

The LACE risk model is used as a comparative benchmark. LACE scored a 0.65 Area Under the Curve (AUC) on the same data and under the same timeframe using an "Adaptive Boosting" model (AdaBoost).

## Chart Notes

Over 99% of all hospitalizations at Providence Health & Services for the period of 2012 through 2014 came with at least one unstructured medical chart note, each containing potentially thousands of words. This represents a large source of additional information that is easily overlooked due to its unstructured (or, at best, semi-structured) format. Note text can capture subtle nuances, such as tone in describing a patient's state of mind, behavior, or appearance [3]. It can fill gaps and corroborate voluntary information given by a patient such as details about lifestyle and smoking history, all pertinent to COPD [4]. As an additional benefit, the data may be largely uncorrelated to structured data in the chart, and therefore may help the predictive power when added to existing models.

## What is *word2vec*?

*word2vec* is an open-source project developed by Tomas Mikolov, Ilya Sutskever, Kai Chen, Greg Corrado and Jeffrey Dean at Google Research in 2013. The original algorithm is programmed in the C language. In this study, we used Radim Rehurek Python implementation of *word2vec* in the Gensim package. The Python port is fast, easy to use, well supported, and comes with thousands of compatible libraries such as the powerful Natural Language Toolkit (NLTK).

One of the examples cited on the first page of *word2vec's* documentation[5], and often mentioned in related articles, describes feeding a portion of the Wikipedia corpus to *word2vec* and asking for the answer to the following analogy question:

*king is to man as what is to woman?*

*word2vec* uses neural networks to create vector representations of words based on their contextual and relational positioning[6]. It creates a multi-dimensional vector representation of each word based on all observed uses. These vectors respond well to basic addition, so the vector representation of 'king -man + woman' will be extremely close to the vector for 'queen'.

*word2vec* works with unlabeled data and handles sentences of any length. Unlike most NLP, stemming and purging of stop words actually harms the model by distorting context[7]. The larger the corpus, the deeper its understanding of the data.

The final *word2vec* product is a matrix of words and arbitrary scores for the number of dimensions selected by the user. For this study, we asked *word2vec* to run through 650 megabytes of data (representing the latest state of all chart notes) and return 500 dimensions for every word encountered at least 100 times (about 284,000 words in all).

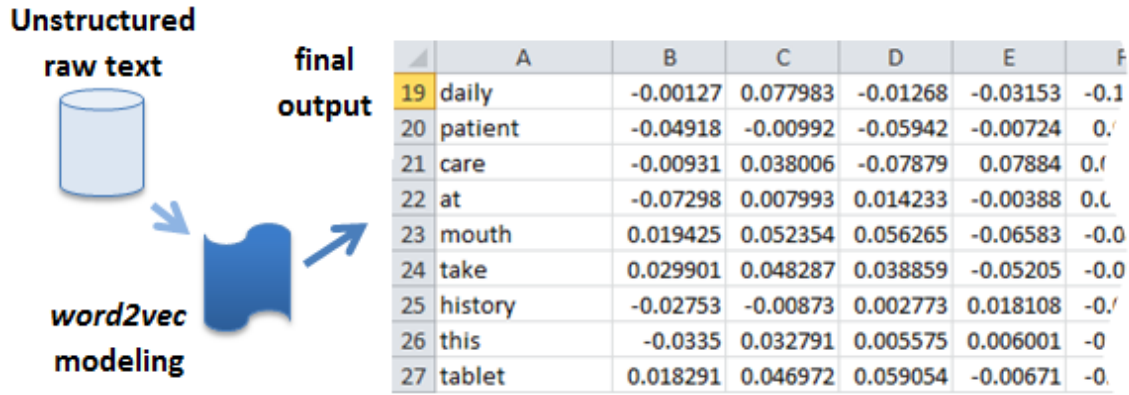

**Two Approaches**

We used two different methods to create predictive variables using *word2vec*:

**1) Supervised Approach: Building Bag-of-Words based on Manually Chosen Terms**
From hand-selected starting words closely related to COPD, we queried *word2vec* to return the most similar words, and then score each encounter's notes based on the amount of content related to those seed words.

**2) Unsupervised Approach: Building Clusters on *word2vec's* output**
As a totally different way to approach the data, we clustered *word2vec's* matrix output into 150 groups and used the clusters to characterize the notes from a hospitalization.

# Method 1: Building Bags-of-Words based on Manually Chosen Terms

## Modeling *word2vec*

a) Pre-processing Text Data
The first step is to train *word2vec* using a text corpus of choice. We used 650 megabytes worth of unique chart notes from COPD patients covering 2012 to mid-2014 (we didn't use any notes from the testing period, though this is a personal choice as *word2vec* doesn't have access to the response variable). We opted to use some data processing to keep things understandable (though this isn't necessarily recommended):

1. Remove numbers.
   2. Remove contiguous spaces.
   3. Remove all punctuations with the exception of periods, semicolons, interrogation marks, and exclamation points.
   4. Create vectors of sentences by splitting each blob of texts on periods, semicolons, interrogation marks, and exclamation points.

b) Building vocabulary
   1. Feed each sentence into *word2vec* to generate the list of all the words meeting the threshold.

c) Modeling
   2. Train *word2vec*.

## Building Bag-of-Words

1. We hand-select words covering different and complementary areas of the COPD story. The terms attempt to be COPD-specific but also descriptors of a patients' situation, environment, personality and state of mind.

| | | | |
|---|---|---|---|
| aclidinium | albuterol | anger | arformoterol |
| asbestos | asthma | bronchodilators | budesonide |
| children | copd | corticosteroids | cough |
| dirty | dizziness | dust | education |
| exacerbation | factory | family | farmer |
| finance | fluticasone | formoterol | friend |
| god | heroin | homeless | indacaterol |
| inflammatory | insurance | ipratropium | irritated |
| levalbuterol | manual | married | medicare |
| methylxanthines | mobility | mold | narcotics |
| noncompliant | partime | parttime | phosphodiesterase |
| pollution | prednisolone | runny | salmeterol |
| scratchy | sick | smoker | support |
| surgery | theophylline | throat | tiotropium |
| tired | tremor | tremors | unemployed |
| vomit | vomiting | welfare | |

2. We then call Gensim's "most_similar" function to gather the 200 closest synonyms for each starting word according to the cosine similarity of the *word2vec* vector representations. Here is an example of the top 20 words for 'asthma' sorted by closeness:

```
In [112]: model.most_similar('asthma',topn=20)
Out[112]:
[('copd', 0.7092980742454529),
 ('ashtma', 0.5638915300369263),
 ('emphysema', 0.5610221028327942),
 ('bronchospasm', 0.5528830289840698),
 ('rhinitis', 0.5293585062026978),
 ('exacerbation', 0.5132066011428833),
 ('astham', 0.5109268426895142),
 ('atopy', 0.5098930597305298),
 ('fibromyalgia', 0.4894115924835205),
 ('asthmaticus', 0.4878374934196472),
 ('sarcoidosis', 0.4817339777946472),
 ('bronchitis', 0.4817228615283966),
 ('rhinits', 0.4717721939086914),
 ('asbestosis', 0.47124117612838745),
 ('asthmatic', 0.46894407272338867),
 ('hypertension', 0.4664614498615265),
 ('obstructive', 0.46565642952919006),
 ('gout', 0.4648151397705078),
 ('gerd', 0.46477368474006653),
 ('arthritis', 0.4536154568195343)]
```

3. We then find all matching words at the bag level and sum the square of the closeness score for each seed word. This total closeness squared value is the final feature used in our COPD model. Because we used 63 starting terms for our bag-of-words, we end up with 63 new features to use in our model.
4. Finally, an AdaBoost model is used to train our new quantitative features on PH&S patients and attempt to predict risk of readmission for COPD patients.

### Modeling and Results

We ran AdaBoost in R (code included) and got an AUC score of 0.63. This is pretty remarkable, as LACE gets 0.65 by pulling the most critical structured elements out of the chart. By hand selecting a series of words, we can easily capture nearly the same amount of information without any assistance from typical EMR data.

## Method 2: Clustering

1) The 160,000 500-dimensional vectors are grouped into 50 clusters based on their vector representation. Because similarity in *word2vec* representations is typically measured by the cosine similarity between vectors, rather than the Euclidean distance, the spherical k-means algorithm is used for this clustering.
2) Each note is scored based on the percentage of its words contained in each cluster. This creates 50 new features (since there are 50 clusters). These columns don't add exactly to 100% since not all words met the 100-count threshold in the corpus, but on average the clusters capture over 96% of the words in the notes.
3) As an alternative measure, an "affinity" to each cluster is computed, based on the sum of the square of each word's cosine similarity to the centroid of its corresponding cluster, across all words in the note. This has the advantage of being sensitive to the volume of notation (likely indicating more serious issues), and distinguishing between words that are central vs peripheral to the main concept of the cluster.

### Modeling and Results
We ran AdaBoost in R (code included) and got an AUC score of 0.61 from the percentage approach and 0.6 from the affinity approach.

### Note on the CMS Definition Used for COPD Patients
Notes and patients were selected using the following CMS definition: An admission with a principal discharge diagnosis indicating acute exacerbation of COPD (ICD-9 of 491.21, 491.22, 491.8, 491.9, 492.8, 493.20, 493.21, 493.22, 496) or both a principal diagnosis of respiratory failure (518.81, 518.82, 518.84, 799.1) and secondary diagnosis of COPD (491.21, 491.22, 493.21, 493.22).

## Conclusion
The speed and ease at which *word2vec* yields quantitative measures from unstructured text is impressive and its potential barely explored in this paper (for example, sentences were treated as bags of words without order, rather than trying to model the effect words have on one another). The challenge is coming up with creative ways to apply this mathematical representation of language to better answer relevant questions.

## Citations

# Appendix

### word2vec Feeder (thanks to Kaggle.com for Python source)

```python
# word2vec sentence cleaner and feeder in Python
# start – assumes various csv files containing one sentence per line in 'notes4word2vec' folder
import csv, re, gensim
from gensim.models import word2vec
foldername = 'notes4word2vec'

# import modules & set up logging
import logging
logging.basicConfig(format='%(asctime)s : %(levelname)s : %(message)s', level=logging.INFO)

# line iterator to feed word2vec and clean each sentence
def getlines(dirname):
    import os
    for fname in os.listdir(dirname):
        for line in open(os.path.join(dirname, fname)):
            yield  re.sub("[^a-zA-Z]"," ", line).split()

# instatiate word2vec model using 500 word vector dimensionality
#     10 context window, 100 minimum word count, 15 workers
model = word2vec.Word2Vec(size=500, window=10, min_count=100, workers=15)

print "Building vocabulary..."
sentences = getlines(foldername) # memory-friendly iterator
model.build_vocab(sentences)

print "Training model..."
sentences = getlines(foldername)
model.train(sentences)

model.save('size500_window10_mincount100_alphaonly_sg1')

# save model as csv file
model.save_word2vec_format('Matrix_size500_window10_mincount100_alphaonly_sg1.csv')
# end
```

### word2vec Feature Maker (Python source)

```python
words2get = ['smoker','family','friend','copd','exacerbation',\
'narcotics','support','finance','dirty','anger','tired', 'education',\
'medicare','welfare','pollution','dust','mold', 'asthma','heroin',\
'insurance','mobility','children','married','homeless','god', 'asbestos',\
'farmer','factory','parttime','partime','surgery','manual', 'unemployed', 'sick',\
'noncompliant', 'tremor', 'tremors', 'dizziness', 'runny', 'irritated', \
'scratchy', 'throat', 'inflammatory', 'vomit', 'vomiting', 'albuterol', \
'levalbuterol', 'ipratropium', 'bronchodilators', 'tiotropium',\
'salmeterol', 'formoterol', 'arformoterol', 'indacaterol', 'aclidinium',\
'corticosteroids', 'fluticasone',  'budesonide', 'prednisolone', 'phosphodiesterase',\
'methylxanthines', 'theophylline', 'cough']

for seedword in words2get:
        print(seedword)
        tt = model.most_similar(seedword,topn=200)
        with open(seedword + '.csv',"w") as thefile:
```

```
            for items in tt:
                    thefile.write(str(items[0]) + "," + str(items[1]) + "\n")
```

## Feature Maker (R source)

```r
library(data.table)
# CleanNotesForMatching has to contain only notes cleaned exactly like word2vec notes
basedf <- data.frame(fread(input = 'CleanNotesForMatching.csv',integer64="numeric"))
basedf_scored <- data.frame('PAT_ENC_CSN_ID'=basedf$PAT_ENC_CSN_ID)

# get all bag of words
filenames <- list.files("copdwordsonly", pattern="*.csv", full.names=TRUE)
for (file in filenames) {
   basedftemp <- basedf['NOTE']

   seachword <- strsplit(strsplit(file, '/')[[1]][2], '\\.')[[1]][1]
   tempdf <- read.csv(file, header=FALSE, stringsAsFactors=FALSE)
   tempdf <- tempdf[1:200,]
   words <- c(seachword,tempdf$V1)
   score <-  c(1,tempdf$V2)
   keywords <- data.frame('seachword'=words, 'distance'=score)

   basedftemp$varnames <- paste(words,sep=" ", collapse=" ")
   temp <- basedftemp[c('NOTE','varnames')]
   counter <- 0
   commondf <- NULL
   for (splt in split(temp, (as.numeric(rownames(temp))-1) %/% 10000)) {
      counter <- counter + 1
      print(counter)
      x <- lapply(splt, strsplit, " ");
      x <- Map(intersect, x[[1]], x[[2]]);
      tempdf <- sapply(x, paste0, collapse = " ")
      if (is.null(commondf)) {
         commondf <-  data.frame('common'=tempdf)
      } else {
         commondf <- rbind(commondf, data.frame('common'=tempdf))
      }
   }

   together <- basedf_scored
   together$common <- as.character(commondf$common)

   together[,seachword] <- sapply(together$common, function(x) {
      sentence <- strsplit(x=x,' ')[[1]]
      sum((keywords[keywords$seachword %in% sentence,]$distance)^2)}
   )

   # aggregate feature by encounter
   word2vecdf <- together[c("PAT_ENC_CSN_ID", seachword)]
   word2vecdf <- aggregate(.~PAT_ENC_CSN_ID, data=word2vecdf, FUN=sum)
   write.csv(word2vecdf, paste0(seachword,'_feature3_square.csv'), row.names=FALSE)
}
```

## Cluster Maker (R source)

```r
require(skmeans)

word2Vec <- read.table('FullCorpus_Matrix.csv',skip=1,row.names=1)

cosineSim <- function(word1,word2) {
```

```r
  x <- as.numeric(word2Vec[word1,])
  y <- as.numeric(word2Vec[word2,])
  x %*% y / sqrt(x%*%x * y%*%y)
}

## Example - similarity between two words
#cosineSim('angry','hostile')

similarWord <- function(word) {
  similarity <- sapply(rownames(word2Vec), function (g) cosineSim(word,g))
  result <- data.frame(word=rownames(word2Vec)[order(similarity,decreasing=TRUE)],
similarity=order(similarity,decreasing=TRUE))
  head(result)
}

## Example - nearest words
#similarWord('angry')

clusterSim <- function(clusterNum, word) {
  x <- clusters$prototypes[clusterNum,]
  y <- as.numeric(word2Vec[word,-ncol(word2Vec)])
  x %*% y / sqrt(x%*%x * y%*%y)
}

clusterRep <- function(clusterNum) {
  similarity <- sapply(rownames(word2Vec)[word2Vec$cluster==clusterNum], function (g) clusterSim(clusterNum,g))
  result <- data.frame(word=rownames(word2Vec)[word2Vec$cluster==clusterNum][order(similarity,decreasing=TRUE)]
, similarity=similarity[order(similarity,decreasing=TRUE)])
  head(result)
}
## Example - word that best captures a cluster
#clusterRep(2)

set.seed(84)
clusters <- skmeans(as.matrix(word2Vec), 150)

word2Vec$cluster <- clusters$cluster

write.csv(data.frame(word=rownames(word2Vec),cluster=word2Vec[,501]),'wordclusters.csv')

clusterSimilarity <- data.frame(word=rownames(word2Vec))
for (i in 1:150) {
    clusterSimilarity[,paste0('cluster',i,'Sim')] <- 0
    clusterSimilarity[word2Vec$cluster==i,paste0('cluster',i,'Sim')] <-
sapply(rownames(word2Vec)[word2Vec$cluster==i], function (g) clusterSim(i,g))
}
write.csv(clusterSimilarity, 'wordClusterSimilarity.csv')
```

## Assign Word Clusters to Notes (R source)
```r
require(tm)
require(reshape2)
# read in notes file with CSN as well as the readmission data
notes <- read.csv('notes.csv', stringsAsFactors=FALSE)
notes$simplified <- stripWhitespace(removePunctuation(removeNumbers(tolower(notes$NOTE_TEXT))))
noteWords <- sapply(notes$simplified, strsplit, split='\\s')
```

```r
noteWords <- unname(noteWords) ##to conserve space

## importing word clusters
clusters <- read.csv('wordclusters.csv',stringsAsFactors=FALSE)
clusterData <- data.frame(csn=numeric(),clusterID=numeric(),propOfNote=numeric())

# Percentage per cluster -------------------------------------------

for (i in 1:length(unique(notes[,1]))) {
 csn <- unique(notes[,1])[i]
 noteRows <- which(notes[,1] == csn)
 wordList <- unlist(noteWords[noteRows])
 totalWds <- length(wordList)
 if (totalWds>0) {
  for (clusterID in unique(clusters$cluster)) {
    wordsInCluster <- clusters$word[which(clusters$cluster==clusterID)]
    temp <- data.frame(csn=csn, clusterID=clusterID, propOfNote=sum(wordList %in% wordsInCluster)/totalWds)
    clusterData <- rbind(clusterData,temp)
  }
 }
 print(paste('admission',i,'of',length(unique(notes[,1])),'complete'))
}
write.csv(clusterData, 'noteClusters_FULL.csv')

clusterData <- read.csv('noteClusters.csv')
clusterData <- subset(clusterData, select=-c(X))
molten <- melt(clusterData,id.vars=c('csn','clusterID'))
columnar <- dcast(molten,formula=csn~clusterID, fun.aggregate= sum )
colnames(columnar) <- c('csn',paste0('cluster',1:150))
write.csv(columnar,'noteClustersColumns_FULL.csv')

# Cumulative  --------------------------------------------------

wordClusterSimilarity <- read.csv('wordClusterSimilarity.csv', row.names=c('word'))
wordClusterSimilarity <- subset(wordClusterSimilarity, select=-c(X))

clusterAddData <- data.frame(csn=numeric(),clusterID=numeric(),affinityOfNote=numeric())
for (i in 1:length(unique(notes[,1]))) {
   csn <- unique(notes[,1])[i]
   noteRows <- which(notes[,1] == csn)
   wordList <- unlist(noteWords[noteRows])
   totalWds <- length(wordList)
   if (totalWds>0) {
      for (clusterID in unique(clusters$cluster)) {
         wordsInCluster <- clusters$word[which(clusters$cluster==clusterID)]
         temp <- data.frame(csn=csn, clusterID=clusterID, affinityOfNote=sum(wordClusterSimilarity[intersect(wordsInCluster,wordList),clusterID]^2))
         clusterAddData <- rbind(clusterAddData,temp)
      }
   }
   print(paste('admission',i,'of',length(unique(notes[,1])),'complete'))
}
write.csv(clusterAddData,'noteClustersAffinity.csv')

molten <- melt(clusterAddData,id.vars=c('csn','clusterID'))
columnar <- dcast(molten,formula=csn~clusterID, fun.aggregate= sum )
colnames(columnar) <- c('csn',paste0('cluster',1:150))
```

```
write.csv(columnar,'noteClustersColumnsAffinity.csv')
```

## AdaBoost Modeler (R source)
```
# model LACE
lace <- read.csv('LACE_COPD.csv')
lace <- lace[,c(setdiff(names(lace),'PAT_ENC_CSN_ID'))]

# response variable true under 30 days
lace$READMITLAG[is.na(lace$READMITLAG )] <- 0
lace$READMITLAG <- ifelse(lace$READMITLAG >= 1 & lace$READMITLAG <=30,1,0)
lace$DISCHARGEDATE <- as.Date(lace$DISCHARGEDATE)

# train/test split data by date
trainObj <- subset(lace, DISCHARGEDATE < "2014-07-01")
testObj <- subset(lace, DISCHARGEDATE >= "2014-07-01")
trainObj <- subset(trainObj, select=-c(DISCHARGEDATE))
testObj <- subset(testObj, select=-c(DISCHARGEDATE))

require(caret)
require(pROC)
trainObj$READMITLAG <- ifelse(trainObj$READMITLAG==1,'yes','nope')
trainObj$READMITLAG <- as.factor(trainObj$READMITLAG)
# train and model
objControl <- trainControl(method='cv',
                           number=10,
                           returnResamp='none',
                           summaryFunction = twoClassSummary,
                           classProbs = TRUE)

objModel <- train(data.frame('LACE'=trainObj$LACE) ,
                           trainObj$READMITLAG,
                           method='ada',
                           metric = "ROC",
                           trControl=objControl)
predictions <- predict(object=objModel, data.frame('LACE'=testObj$LACE), type='prob')
auc <- roc(testObj$READMITLAG, predictions[[2]])
print(auc$auc) # 0.6528

# model all closeness sums of all notes
lace <- read.csv('LACE_COPD.csv')
outcomeName <- 'READMITLAG'
lace$READMITLAG[is.na(lace$READMITLAG )] <- 0
lace$READMITLAG <- ifelse(lace$READMITLAG >= 1 & lace$READMITLAG <=30,1,0)
 lace$DISCHARGEDATE <- as.Date(lace$DISCHARGEDATE)

filenames <- list.files("word2vecSeedBagOfWords_Top200", pattern="*.csv", full.names=TRUE)
fullset <- NULL
for (file in filenames) {
   print(file)
   tempdf <- read.csv(file, stringsAsFactors=FALSE)

   if (is.null(fullset)){
      fullset <- tempdf[,c('PAT_ENC_CSN_ID',names(tempdf)[ncol(tempdf)]]]
   } else {
      tempdf <- tempdf[,c('PAT_ENC_CSN_ID',colnames(tempdf)[ncol(tempdf)]]]
      fullset <- merge(x=fullset, y=tempdf, by='PAT_ENC_CSN_ID', all=TRUE)
   }
}
# merge back with LACE to get discharge date
```

```r
alltogether <- merge(lace, fullset, by='PAT_ENC_CSN_ID')
alltogether <- subset(alltogether, select=-c(PAT_ENC_CSN_ID, LACE))
alltogether$DISCHARGEDATE <- as.Date(alltogether$DISCHARGEDATE)
alltogether$READMITLAG <- ifelse(alltogether$READMITLAG==1,'yes','nope')
alltogether$READMITLAG <- as.factor(alltogether$READMITLAG)
trainObj <- subset(alltogether, DISCHARGEDATE < "2014-07-01")
testObj <- subset(alltogether, DISCHARGEDATE >= "2014-07-01")
trainObj <- subset(trainObj, select=-c(DISCHARGEDATE))
testObj <- subset(testObj, select=-c(DISCHARGEDATE))
trainObj <- na.omit(trainObj)
predictorNames <- setdiff(names(trainObj), outcomeName)
# train and model
objControl <- trainControl(method='cv',
                           number=10,
                           returnResamp='none',
                           summaryFunction = twoClassSummary,
                           classProbs = TRUE)
objModel <- train(trainObj[,predictorNames],
                  trainObj[,outcomeName],
                  method='ada',
                  metric = "ROC",
                  trControl=objControl)
predictions <- predict(object=objModel, testObj[,predictorNames], type='prob')
auc <- roc(testObj$READMITLAG, predictions[[2]])
print(auc$auc) # 0.6282
```